\documentclass{article}
\usepackage{kr}

\usepackage{times}
\usepackage{soul}
\usepackage{url}
\usepackage[hidelinks]{hyperref}
\usepackage[utf8]{inputenc}
\usepackage[small]{caption}
\usepackage{graphicx}
\usepackage{amsmath}
\usepackage{amsthm}
\usepackage{booktabs}
\usepackage{algorithm}
\usepackage[noend]{algpseudocode}
\usepackage{todonotes}

\theoremstyle{plain}

\newtheorem{theorem}{Theorem}
\theoremstyle{definition}
\newtheorem*{example}{Example}

\title{Planning Domain Model Acquisition from State Traces without Action Parameters}
\author{%
Tom\'a\v{s} Balyo$^1$\and
Martin Suda$^{1,2}$\and
Luk\'a\v{s} Chrpa$^{1,2}$\and 
Dominik \v{S}afr\'anek$^{1,3}$\and \\
Stephan Gocht$^1$\and
Filip Dvo\v{r}\'ak$^1$\and
Roman Bart\'ak$^{1,4}$\and
G. Michael Youngblood$^1$ \\
\affiliations
$^1$Filuta AI, Inc., 1606 Headway Cir STE 9145, Austin, TX 78754, United States\\
$^2$Czech Institute of Informatics, Robotics and Cybernetics, Czech Technical University, Prague, Czechia\\
$^3$Center for Theoretical Physics of Complex Systems, Institute for Basic Science, Daejeon, Korea\\
$^4$Faculty of Mathematics and Physics, Charles University, Prague, Czechia\\
\emails
tomas@filuta.ai, martin.suda@cvut.cz, lukas.chrpa@cvut.cz, dsafranek@filuta.ai,\\
stephan@drgocht.com, filip@filuta.ai, bartak@ktiml.mff.cuni.cz, michael@filuta.ai
}

\begin{document}

\maketitle

\begin{abstract}
Existing planning action domain model acquisition approaches consider different types of state traces from which they learn. The differences in state traces refer to the level of observability of state changes (from full to none) and whether the observations have some noise (the state changes might be inaccurately logged). However, to the best of our knowledge, all the existing approaches consider state traces in which each state change corresponds to an action specified by its name and all its parameters (all objects that are relevant to the action). Furthermore, the names and types of all the parameters of the actions to be learned are given. These assumptions are too strong.

In this paper, we propose a method that learns action schema from state traces with fully observable state changes but without the parameters of actions responsible for the state changes (only action names are part of the state traces). Although we can easily deduce the number (and names) of the actions that will be in the learned domain model, we still need to deduce the number and types of the parameters of each action alongside its precondition and effects. We show that this task is at least as hard as graph isomorphism. However, our experimental evaluation on a large collection of IPC benchmarks shows that our approach is still practical as the number of required parameters is usually small.

Compared to the state-of-the-art learning tools SAM and Extended SAM
our new algorithm is able to provide better results in multiple domains in terms of learning action models more similar to reference models, even though it uses less information and has fewer restrictions on the input traces.
%
\end{abstract}

\section{Introduction}

Automated Planning is an important field of Artificial Intelligence that deals with the problem of finding a sequence of actions, i.e., a plan, that transforms the state of the world from the initial state to some goal state. Domain-independent planning requires a symbolic model describing the world and actions that is fed to a planning engine, a generic solver of planning tasks. Such a symbolic model can be represented, for instance, in the well-known PDDL language~\cite{pddl}.

Domain model acquisition, in a nutshell, deals with the problem of acquiring such a symbolic model that adequately represents a given domain. Tools such as GIPO~\cite{simpson2007planning}, ItSimple~\cite{vaquero2007itsimple}, or, recently, \texttt{planning.domains} provide very useful support in domain modeling. These tools, however, still rely on an expert who handcrafts the model. 

Automated domain model acquisition aims at mitigating or even alleviating the need for an expert to encode symbolic domain models. The underlying idea behind automating domain model acquisition is to leverage observations of world states (and their changes) and/or of actions that an observed entity performs. In other words, data from which the domain acquisition tools synthesize domain models are in the form of traces of states and actions. The tools differ by ``completeness'' of traces that they support, ranging from complete traces (full states and actions) in the case of Observer~\cite{observer} or SAM~\cite{sam}, noisy traces (some information is missing or is incorrect) in the case of ARMS~\cite{arms} or FAMA~\cite{fama}, or traces containing only actions in the case of LOCM~\cite{locm}. However, to the best of our knowledge existing domain acquisition tools from state traces require full information about the parameters of the action to be learned. Although LSSS~\cite{DBLP:conf/ecai/BonetG20} does not require action parameters either, it, however, requires the whole reachable state space for each training instance (which might be difficult to obtain and huge).

The assumption that we know (all) action parameters is too strong. For example, we might want to learn a domain model of a game from state traces that are logged from the gameplay~\cite{DBLP:journals/corr/abs-2402-12393}. Although we can assume that we have full observability of states (the game mechanics has that information albeit the player might not see everything), we might have limited information about the actions that the player took. Usually, we can extract the name of the action that the player took (e.g., move, pickup). However, we might not be able to extract correct and complete sets of parameters of the actions as the implementation of the game mechanics might consider some parameters implicit. For example, the location of an item the player wants to pickup might not be explicitly required in the game implementation, yet it is important for the accurate description of the (planning) action since the player must be at the same location as the item. That means that state traces might not contain all action parameters (if any) and hence it is up to the domain model acquisition method to infer action parameters (alongside their preconditions and effects) from such state traces.

In this paper, we address this gap in the literature by proposing 
a new algorithm for domain model acquisition that operates 
over state traces that contain only action names and assume no information about the action arguments.
In other words, the information about which objects get substituted for the respective action's parameters in each step is missing.

The introduced algorithm leverages the idea of generating sets of possible explanations that, in a nutshell, have to explain that a synthesised action schema correctly handles all transitions (between states) in the traces in which the action is involved. We have empirically evaluated our algorithm---on a range of IPC domains---in terms of quality (how the synthesised models differ from original IPC domains). We have also compared our algorithm to the state-of-the-art domain acquisition tools SAM and Extended SAT~\cite{sam} and the results show that our algorithms outperform both SAM versions in quality for multiple planning domains.

\section{Preliminaries}
The planning problem deals with finding a plan, a sequence of grounded actions, that transforms the world from an initial state to some goal state, i.e., a state where all goal conditions are satisfied.

A (first-order-logic) \emph{predicate} is represented by its unique name and a set of $k \geq 0$ arguments. A lifted (resp.~grounded) predicate has variables (resp.~objects) as its arguments. We consider a \emph{typed} representation, i.e., each variable (resp.~object) has a certain type, which means that only objects of a given type can be substituted for a variable of that type. A \emph{state} is represented by a set of grounded predicates, or \emph{facts}, that are understood to be true in that state. 

An \emph{action} $a$, spelled out as $\mathit{act\_name}(x_1,\dots,x_k)$, consists of a unique name $\mathit{act\_name}$ and a (possibly empty) list of distinct variable arguments (often referred to as action's \emph{parameters}) $x_1,\dots,x_k$, and is determined by three sets of (lifted)
predicates $\mathit{pre}(a)$, $\mathit{add}(a)$ and $\mathit{del}(a)$, 
representing $a$'s preconditions, add and delete effects, respectively.
The predicates in $\mathit{pre}(a)$, $\mathit{add}(a)$ and $\mathit{del}(a)$
are over the variables $x_1,\dots,x_k$. 
Such a lifted action
(we sometimes say \emph{action schema} to stress the presence of variables)
is grounded (we may also say \emph{instantiated}) by providing a \emph{substitution},
a mapping $\sigma$ from the variables $x_1,\dots,x_k$ to (not necessarily distinct) objects $o_1,\dots,o_k$,
and consistently grounding the predicates in $a$'s preconditions and effects using $\sigma$.

A grounded action $a$ is \emph{applicable} in a given state $S$ if and only if $\mathit{pre}(a)\subseteq S$. A grounded action that is applicable in a state $S$ transforms the state $S$ to a new state $S'$ such that $S'=(S\setminus\mathit{del}(a))\cup\mathit{add}(a)$, we also write $S'=\mathit{apply}(S,a)$.

A \emph{planning problem instance} consists of a \emph{domain definition} and a \emph{problem definition}. The domain definition consists of a set of predicates, a set of lifted actions, and a hierarchy of types. 
The problem definition consists of the set of objects with their types and two sets of grounded predicates, one representing the initial state $I$ and the other the goal conditions $G$. A \emph{plan} for a given problem instance is a sequence of grounded actions $\pi=\langle a_1,\dots,a_n\rangle$ such that
$G\subseteq \mathit{apply}(\dots\mathit{apply}(\mathit{apply}(I,a_1),a_2)\dots a_n)$ meaning the consecutive application of actions in $\pi$, when starting in the initial state, is possible and results in a state in which the goal conditions are satisfied. 

In general, a \emph{state trace} is an alternating sequence of states and grounded actions 
$S_0, a_1, S_1, \dots, a_n, S_n$, where $a_i$ is applicable in $S_{i-1}$ and $a_i$ transforms $S_{i-1}$ to $S_i$. We consider fully observable states $S_i$ and (only) names of the ground actions $a_i$. Note that in literature, the states might be partially observable and/or noisy.


\emph{Domain Model Acquisition} (or domain synthesis) is the task of generating a domain definition file from a set of traces. That is, in our case, deducing the set of lifted actions $\mathcal{A}$ such that for each pair of subsequent states $S_{i-1}$ and $S_{i}$ and the action name $act\_name_i$ in between, there is a grounded action $a$, an instance of some lifted action from $\mathcal{A}$, such that $\mathit{act\_name}_i$ is the name of $a$, $a$ is applicable in $S_{i-1}$, and $S_i=\mathit{apply}(S_{i-1},a)$.

%
%
%
%
%
%

\section{Related Work}

Automated domain synthesis is a field that originated in the early 90s. Since then, numerous methods for automated action model acquisition have been developed, some of which have been implemented in the MACQ framework~\cite{macq}. The most relevant that are related to ours can be collected into six groups based on their overall similarity and the set of working assumptions. Almost all domain acquisition methods assume the STRIPS hypothesis space with a few exceptions, which we mention below. Similarly, all but one acquisition model assumes deterministic effects.

The first group are the earliest versions of model acquisition techniques which assume full observability of the world states, without any noise. Considering that observed states are trustworthy, these models operate on full satisfiability. The learning paradigm is based on interaction with the environment: the models learn by successful and unsuccessful execution of the plans, with occasional and optional input from the expert. The prominent examples include OBSERVER~\cite{observer}, which assumes fully observable traces, which are strings of state, action, state, action, etc., where the first and the final state describe the initial and the goal states. Actions have labels, but they may have preconditions and effects partially or completely missing. The paradigm of learning preconditions is based on having two competing sets: those that overapproximated the set of preconditions and those that underapproximate it. In ideal conditions, as the model learns by failing to execute plans, these two converge to each other. There are some limitations, however. While the model allows negated preconditions, it cannot learn them, in which case it requires input from the human expert. OpMaker~\cite{DBLP:conf/aips/McCluskeyRS02}, designed as a part of GIPO, can be understood as an automated ``assistant'' for developing action models that share a lot of similarities with OBSERVER while supporting human user intervention. Opmaker also supports the acquisition of hierarchical action models. Another example is EXPO~\cite{expo} which operates with the same set of assumptions as OBSERVER. While it uses STRIPS, it also optionally allows conditional effects. Here, the learning is based on the Operator Refinement method. It learns missing preconditions from unpredicted outcomes, and incomplete effects from unpredicted states, when executing plans. TRAIL~\cite{TRAIL}, which also uses fully observed states, is loosely related. However, instead of STRIPS, the action model uses action-effect pairs, called Teleo-operators, which can be applied as long as necessary to achieve the intended effect. TRAIL uses Inductive Logic Programming to learn the preconditions for the action models. OLAM learns STRIPS action models incrementally (from fully observable traces) by effectively interleaving action model learning and exploration phase selecting plans for execution~\cite{DBLP:conf/ijcai/LamannaSSGT21}. Recent work~\cite{DBLP:conf/aaai/BachorB24} provides a theoretical and complexity analysis of domain acquisition under the assumption of ``justified'' state traces (i.e., actions performed in state traces are not redundant). It has been shown that it is sufficient to consider the most restrictive models, maximizing preconditions and delete effects while minimizing add effects.

The second group operates with only partially observed or noisy states. Because of this, the learning paradigm of action preconditions and effects is based on optimization, rather than satisfaction. The goal is to create the most likely action model, given some metrics. The first such model and the former state-of-the-art algorithm is ARMS~\cite{arms}. In ARMS, the initial and final states are known, but intermediate states may be incomplete, and the fluents are hidden. The learning is framed as an optimization task that minimizes the error and redundancy metrics, aiming to find concise and correct models. It does that by creating a weighted propositional satisfiability problem and solving it with MAX-SAT. The first of the two extensions of this method is LAMP~\cite{LAMP}, which additionally includes learning action models with universal and existential quantifiers and logical implications, using ADL instead of the STRIPS model hypothesis. Additionally, it uses the Markov Logic Network for the optimization. The second is AMAN~\cite{aman}, which extends ARMS to include situations in which actions have a probability of being observed incorrectly and uses the same MAX-SAT. AMAN has been further extended as ADMN~\cite{admn}, to include noisy states and disordered or parallel actions. Loosely related are LOUGA~\cite{louga}, ALICE~\cite{alice}, and SLAF~\cite{slaf} methods. They use the same input but approach the problem differently. LOUGA uses a genetic algorithm, in which the learning is based on evolving the model to achieve maximal fitness. The fitness function is defined in terms of preconditions not met by the action model, redundant effects, and the number of state variables missing in the trajectory. ALICE uses triples (pre-state, action, post-state) instead of traces and assumes that state observation might be noisy, instead of missing. The learning is based on Boolean classifiers. First, the effects are learned from the parameters of these classifiers, and preconditions are derived later in the post-processing stage. Finally, SLAF assumes a fully observable sequence of executed actions, and partially observed states, but without noise. It uses logical encoding to extract action theories based on the constraint propagation approach, by which it differs from other, optimization-based approaches. Given observations of actions and partially observed states, inconsistent models are removed, and only those action models that are consistent with the observations remain. Recent work concerns AMLSI (Action Model Learning with State machine Interaction) family of domain acquisition techniques that learn STRIPS, temporal, and hierarchical action models~\cite{DBLP:phd/hal/Grand22}.

The third group models the planning as a labeled directed graph, where edges are labeled with action names. LOCM~\cite{locm} assumes a sequence of executed actions but no knowledge of the actual states. Actions contain parameters. It learns predicates, including static predicates, and preconditions and effects of actions, by using a finite state machine. The lack of observed states is compensated by reasonable assumptions in the structure of actions. The two follow-ups include LOCM2~\cite{locm2}, which includes multiple finite state machines, and LOP~\cite{lop}, including preconditions with static predicates. Then, LSSS~\cite{DBLP:conf/ecai/BonetG20} explores reachable state spaces of problem instances without the explicit knowledge of predicates, only with labels denoting types of actions. It learns the predicates of the operators and the initial state by using the satisfiability paradigm with a two-level combinatorial search. The outer level learns the simplest values of the parameters, and the inner level computes the action schemes consistent with the observations.

The fourth group contains LPROB~\cite{lprob}, which is the only method that assumes non-deterministic action effects. The state traces (pre-state, action, post-state) are fully observed. It uses Probability STRIPS and Prob SAT optimization scheme, to learn preconditions and effects, and probabilities associated with the non-deterministic action effects.

The fifth group contains only one the FAMA method \cite{fama} that uses only a minimal set of assumptions: the initial state and goal. It can use incomplete sequences of observations---an unbounded number of unobserved states and actions may be missing. However, it is assumed there is no noise in the traces. It uses STRIPS but allows conditional effects. Here, the lack of observed actions is compensated by compiling the problem into a planning scheme, which allows the planner to fill the gaps consistent with the observations.

Finally, the sixth group consists of a family of methods that follows the concept of \emph{safe model-free planning}~\cite{DBLP:conf/ijcai/SternJ17}, which, in a nutshell, guarantees that a plan generated from state traces will not fail. \emph{Safe Action Model (SAM)} learning leverages this concept for acquisition of lifted domain models~\cite{sam}. Initially, two versions of SAM were proposed. The first version assumes that each object is only ever bound to one action parameter at a time in the state traces (called the injective action binding assumption) and if the number of predicate arguments is bounded, it is sufficient to observe a linear number of state traces with respect to the size of the lifted action model. The second version of SAM relaxes the injective binding assumption of the first version~\cite{sam}. The SAM approach has been extended to learning action models with numeric fluents~\cite{DBLP:conf/aaai/MordochJS23} and probabilistic action models~\cite{DBLP:conf/aaai/JubaS22}. Most recently, the SAM approach has been extended to deal with partially observable state traces~\cite{DBLP:conf/aaai/LeJS24}.

\section{Domain Model Acquisition}

From the user's perspective, domain model acquisition starts from some sort of logs
collected while interacting with an environment of interest. In our case, such a log
records a sequence of states and with each state the set of ground predicates that become true or false since the previous state, i.e., their truth status changed from false to true, resp.~from true to false. For each state, we additionally record a single label that represents the name of the action responsible for transforming the previous state to the current state. States with the same label belong together and will be synthesized into a single planning action with its name equal to the label.

We start this section by spelling out the exact input and output requirements for our domain model acquisition task, and, in particular, by showing how the informal idea of logs gets captured by the notions of traces and transitions.

\subsection{Task}

As the input for a domain model acquisition task, we consider
\begin{itemize}
    \item a hierarchy of PDDL types $(\mathcal{T},\leq)$, where $\leq$ is the ``subtype''  relation; we expect $\top \in \mathcal{T}$ such that we have $t \leq \top$ for any $t \in \mathcal{T}$ ($\top$ is the PDDL \emph{object} type),
\item the set of predicates that we expect to encounter $\mathcal{P}$. We get their names and typing information; e.g., $p : t_1 \times \ldots \times t_k$ where $k$ is the arity of $p \in \mathcal{P}$ and $t_i \in \mathcal{T}$,
\item a set of traces with states over grounded predicates from $\mathcal{P}$ using some finite set of objects $\mathit{Obj}$. Each such trace is of the form $S_0, a_1, S_1, \dots, a_n, S_n$, however, we do not have access to the actions $a_i$ themselves, only to their respective names (which we will also call action \emph{labels}). So we actually work with traces of the form:
$S_0, (l_1=\mathit{name}(a_1)), S_1, \dots, (l_n=\mathit{name}(a_n)), S_n$.
\end{itemize}

Note that the order in which the states appear in the traces is for us only relevant 
for capturing the immediate neighbouring states and the action label in between. Thus the given traces naturally decompose into a single set of \emph{transitions} $\mathcal{S}$.
This allows us to combine traces from multiple problem instances and even partial traces
(traces where we do not start in the initial state and/or the goal of the problem instance is not achieved).
Each transition $R \in \mathcal{S}$ is a triple $(S_b,l,S_a)$, where the \emph{before}-state $S_b$ and \emph{after}-state $S_a$ are sets of ground facts, each fact being a known predicate $p\in \mathcal{P}$ applied to a vector of objects $\vec{o}$, the length of which matches the arity of $p$.
Moreover, $l$ is an action label from some finite set of labels/names $l\in \mathcal{L}$.

It will be natural for us to treat each transition in isolation or to group them together according to the labels $l$. The remaining temporal information contained in the traces can be safely discarded.



As the output, we want to synthesise, for every action label $l$ appearing in the given transitions $\mathcal{S}$ a lifted action (schema) $a_l = (\vec{r},\vec{t},\mathit{pre},\mathit{add},\mathit{del})$ consisting of 
\begin{itemize}
    \item a vector of distinct parameters $\vec{r}=r_1,\ldots,r_k$ (of yet-to-be-determined length $k$),
    \item typing information $\vec{t}=t_1,\ldots,t_k$ (a vector over $\mathcal{T}$),
    \item a set of preconditions $\mathit{pre}$, add effects $\mathit{add}$ and delete effects $\mathit{del}$, each of which is a set of (lifted) facts, 
    i.e., predicates $p\in \mathcal{P}$ applied to some of our parameters from $\vec{r}$ (possibly with repetitions, in an arity-respecting and typing-respecting manner),
\end{itemize}
such that (\emph{explains observed transitions}):
for every transition $R = (S_b,l,S_a) \in \mathcal{S}$, 
there is a (types-respecting\footnote{I.e., the object $\sigma_R(r_i)$ can only be of type $t_i$ (or more specific).}) substitution $\sigma_R: \{r_1,\ldots,r_k\} \to \mathit{Obj}$
satisfying
\begin{itemize}
    \item 
    $f\sigma_R \in S_b$ for every $f\in \mathit{pre}$, and
    \item
    $S_a = (S_b \setminus \{e\sigma_R \mid e \in \mathit{del}\}) \cup  \{e\sigma_R \mid e \in \mathit{add}\}.$
\end{itemize}

%

We will refer to the problem specified above as \emph{action model synthesis}.


\subsection{Complexity}

We show that the problem of synthesising an action with $k$ arguments is at least as hard as the graph isomorphism problem and is contained in NP.

\begin{theorem}
The following problem is in NP: Given a set of transitions and a number $k$, is there a lifted action with $k$ arguments and a substitution for each transition that grounds the action consistently ?
\end{theorem}
\begin{proof}
We have to prove that we can check a given witness (a lifted action and a substitution for each transition) in polynomial time. For each transition, we apply the given substitution to the lifted action to obtain a ground action. To check if the ground action is consistent, we need to confirm that: the preconditions are satisfied, there is an effect in the ground action for every difference between the before and after state of the transition and each effect is true in the after state. All of these checks can be performed via set containment operations in polynomial time. 
\end{proof}

\begin{theorem}
The following is graph isomorphism hard: Given a set of transitions and a number $k$, find a lifted action with $k$ arguments and substitutions that ground the action consistently with the transitions.
\end{theorem}

\newcommand{\edgepredicate}[0]{\mathit{edge}}

\begin{proof}
We encode a graph isomorphism problem into action model synthesis. Let $G_1 = (V_1, E_1)$, $G_2 = (V_2, E_2)$ be two graphs and we want to find a function $f: V_1 \mapsto V_2$, showing that the graphs are isomorphic. Meaning that $f$ is a surjective and injective function that maps nodes from the first graph to the second, such that $(v_1, v_2) \in E_1$ if and only if $(f(v_1), f(v_2)) \in E_2$. Note that we can assume that there are no isolated vertices, i.e., vertices not incident to any edges. If such vertices exist then the graphs are only isomorphic if both graphs have the same number of isolated vertices, which can be mapped arbitrarily as there are no edges. Hence, we only need to consider the graph isomorphism for the subgraphs that do not contain isolated vertices. Another edge case to consider is that the graphs have a different number of nodes, in which case no isomorphism exists. Hence we assume that both graphs have the same number of nodes. 

We construct a domain synthesis problem to encode the graph isomorphism problem. There is an object for each node in graph $V_1$ and $V_2$ respectively, the objects are disjoint. There is a binary predicate $\edgepredicate$. The initial state has all predicates set to false. In the first transition $\edgepredicate(v, v')$ becomes true for each $(v, v') \in E_1$. In the second transition $\edgepredicate(v, v')$ becomes true for each $(v, v') \in E_2$.

``$\Rightarrow$'' If the graphs are isomorphic, observed by the function $f$, then we can construct an action for the domain synthesis problem: The action has a parameter $p_i$ for each node in $v_i \in V_1$. Let $\sigma_1(p_i) = v_i$ and $\sigma_2 = f \circ \sigma_1$. The action has add effects $\edgepredicate(\sigma^{-1}_1(v), \sigma^{-1}_1(v'))$ for each $(v, v') \in E_1$. By construction, $\sigma_1$ grounds the action consistently with the first transition. Furthermore, $\sigma_2$ grounds the action consistently with the second transition, because $f$ demonstrates the graph isomorphism.

``$\Leftarrow$'' Assume we synthesize a lifted action that has $k = |V_1| = |V_2|$ arguments and let $\sigma_1$ be the substitution that maps action arguments to objects so that the lifted action is ground consistently with the first transition. Firstly, note that the substitution $\sigma_1$ must be a bijection: The number of arguments $k$ is equal to the number of objects that appear in predicates changing the state. The number of such objects must be exactly $|V_1|$, because we ensured that there are no isolated vertices and hence there is at least one added edge predicate for each vertex. Secondly, the lifted action found encodes a graph $G_3 = (V_3, E_3)$, where the parameters correspond to nodes and add effects correspond to edges. And finally, this graph must be isomorphic to $G_1$, where $\sigma_1$ maps the nodes from $G_3$ to $G_1$: If $(v, v') \in E_3$, i.e., the lifted action has an add effect $\edgepredicate(v, v')$
then there must be a ground predicate $\edgepredicate(\sigma_1(v), \sigma_1(v'))$ in the after state~$S_a$, which means that, by construction, there is an edge $(\sigma_1(v), \sigma_1(v')) \in E_1$. If there is an edge $(v, v') \in E_1$ then, by construction, there is a predicate $\edgepredicate(v, v')$ in the transition that must be explained by the action and thus $\edgepredicate(\sigma_1^{-1}(v), \sigma_1^{-1}(v'))$ must be an effect, i.e., $(\sigma_1^{-1}(v), \sigma_1^{-1}(v')) \in E_3$. Analogously, $G_3$ is isomorphic with $G_2$ and hence, $G_1$ and $G_2$ are also isomorphic shown by $f = \sigma_2 \circ \sigma_1^{-1}$, where $\sigma_2$ is the substitution grounding the lifted action consistently with the second transition.
\end{proof}

\subsection{Algorithm L1}

Approaching the action model synthesis problem with certain completeness guarantees in mind requires us to 
potentially speculate all possible effects that would explain the observed differences between the before-states and the after-states. We might then check for each such speculated explanation whether it is consistent with every observed transition for its respective action label. 
A challenge to this approach comes from the fact that, in each transition, we also need to speculate over the exact form of $\sigma_R$, the substitution mapping of the action's formal parameters to concrete objects. What is worse is that at the beginning we do not even know the number of parameters each action could possibly have.

\subsubsection{Substitutions and Effects}

Let us start by explaining how we address this last challenge. At the beginning, for each action label $l$ we scan all the transitions belonging to $l$ and in each count the number of distinct objects that occur as arguments to facts that change value there. We denote by $\mathit{min\_pars}_l$ the maximum over these numbers for $l$. It is a lower bound for the number of parameters of the action we will synthesise for $l$. We use $\mathit{min\_pars}_l$ as the initial number of $l$'s parameters. Should it later become impossible to synthesise the action for $l$ with this many parameters, we would increase the value by 1 and start the synthesis (for that particular $l$) from scratch. 
\begin{example} To see why $\mathit{min\_pars}_l$ might be too low in general, consider the following two transitions: 
\[R_1=(\{\},l,\{p(a)\}), R_2=(\{p(a),p(b)\},l,\{p(b)\}).\]
While only one object (namely $a$) is always mentioned as an argument to a predicate of a fact that changes value, it is clear that a single-parameter action cannot explain
both $R_1$ and $R_2$ as one of the transitions requires an addition and the other a deletion to explain, which need to be distinct.
The two transitions, however, can be explained by an action schema with two parameters $x_1,x_2$,
an add effect $p(x_1)$, a delete effect $p(x_2)$ and substitutions 
\[\sigma_{R_1} = \{x_1 \mapsto a, x_2 \mapsto b\},\sigma_{R_2} = \{x_1 \mapsto b, x_2 \mapsto a\}.\]
\end{example}
We remark that in our experiments with the IPC benchmarks, we never arrived at a final number of action's parameters larger than $\mathit{min\_pars}_l$.


The second challenge, after the number of action parameters is fixed, is to speculate the shape of $\sigma_R$ for each transition $R$. Because this leads to a problem where decisions should be tried out to be possibly retracted later, we choose an approach based on encoding the whole remaining synthesis problem into propositional satisfiability (SAT) to outsource the burden of backtracking.

To maintain efficiency, the encoding is incremental, starting from a single transition $R = (S_b,l,S_a)$ and from only encoding the part that makes sure all the observed changes are explained.
When this is done for a single transition, the for-now-preliminary solution always correctly explains it. Let us now introduce this first part of the encoding before we elaborate on what needs to be added to correctly cover all the transitions.

We work with two basic sets of propositional variables.\footnote{Remaining variables are auxiliary, serving to build a compact encoding, but otherwise uninteresting.} The first set describes the shape of $\sigma_R$ in an obvious way: for every formal parameter $p_i$ we have the variables $\mathit{bind}(R,p_i,o_j)$, where $o_j$ ranges over all the objects $\mathit{Obj}$ (of the type prescribed to $p_i$) and encodes that $\sigma_R(p_i)=o_j$.\footnote{The $R$ here in $\mathit{bind}(R,p_i,o_j)$ should be understood as a unique identifier of the transition carrying no other information.} We state in our encoding that exactly one $\mathit{bind}(R,p_i,o_j)$ should be true (for every $p_i$ and for every $R$ from our logs with label $l$).
%
%
The second set contains variables of the form $\mathit{add}(\mathit{pred},p_i,\ldots)$ (and $\mathit{del}(\mathit{pred},p_i,\ldots)$) for every predicate $\mathit{pred}$ among the facts in $S_a \setminus S_b$ (resp. $S_b \setminus S_a$) and every vector of formal parameters $p_i,\ldots$ of a length corresponding to the arity of $\mathit{pred}$. Note that unlike the $\mathit{bind}$ variables these are not indexed by $R$ as their meaning is for the actual action schema (i.e., it is non-local).

We explain, in the encoding, e.g., an addition of $\mathit{pred}(a,b)$ observed in $R$ by adding the (CNF of)
\begin{align}
\label{encode_explain}
\bigvee \mathit{add}(\mathit{pred},p_i,p_j) \land \mathit{bind}(R,p_i,a) \land \mathit{bind}(R,p_j,b),
\end{align}
where the big disjunction is over all the possible vectors of (available) formal parameters of length two.

If the ensuing formula is satisfied minimally with respect to the positive literals over $\mathit{add}(\mathit{pred},p_i,\ldots)$ and $\mathit{del}(\mathit{pred},p_i,\ldots)$,
the corresponding solution will correctly explain our transition $R=(S_b,l,S_a)$. To get a minimal solution we employ a standard trick using \emph{unit literal assumptions}~\cite{minisat}: We consider a literal that is set to true and check if the formula can be satisfied with this variable set to false, until no more literals can be set to false.

While we will get an explanation for the considered transition, the found solution may fail to explain other transitions for $l$ or it may fail to be consistent with non-changes (adding something which is not seen to have been added, or deleting something not seen to have been deleted). In principle, we could add \eqref{encode_explain} for every transition for $l$ into one big formula to prevent the first problem. The formula might, however, become prohibitively large for large input logs. Instead, we now set out to independently, with each transition, \emph{verify} the current solution (as encoded in the assignment to the $\mathit{add}$ and $\mathit{del}$ variables). Only if the verification fails for a particular transition $R'$, we start, one by one, forming bigger and bigger sets of transitions which need to be explained jointly. (E.g., after the first failure, we would go and explain jointly for $\{R,R'\}$). The need for a joint encoding of more than one transition, however, arises only very rarely in the IPC benchmark synthesis.


The difference between synthesis and verification is that with verification we already have a tentative solution (in the form of an assignment to the $\mathit{add}$ and $\mathit{del}$ variables), we supply it to a SAT solver as unit assumptions and thus only ask it to solve for the $\mathit{bind}$ variables, effectively guessing the substitution $\sigma_R$ for a particular $R$.

There is still a missing part of our encoding needed both for verification and for synthesis over larger transition sets than size 1 (as explained above). This is the part that ensures consistency of the speculated effects with non-changes. In our algorithm, we only add formulas of this part in a lazy, on-demand, fashion, always only after a certain $\sigma_R$ has been recovered from a preliminary solution. The trigger for this is seeing an effect, e.g., a true variable $\mathit{add}(pred,p_1,p_2)$ in the tentative assignment, along with, e.g., $\mathit{bind}(R,p_1,a)$ and $\mathit{bind}(R,p_2,b)$ set true as well, while the corresponding grounded $p(a,b)$ is not in $S_a$ for $R$. 

The remedy is to exclude such a solution by adding the following clause to the encoding
\begin{align}
\label{encode_consistent}
\neg \mathit{add}(\mathit{pred},p_1,p_2) \lor \neg \mathit{bind}(R,p_1,a) \lor \neg \mathit{bind}(R,p_2,b)
\end{align}
or analogously for an unobserved delete effect\footnote{Strictly speaking, with delete effects we should speculate the unobserved deletion is possibly ``shadowed'' by an add effect which locally binds to the same ground fact. (This asymmetry comes from the adopted action semantics, where the delete effects are applied before the add effects, as in our $S'=(S\setminus\mathit{del}(a))\cup\mathit{add}(a)$.)
We left out this part from our encoding for simplicity and never encountered an IPC-derived synthesis problem where this omission would lead to an unsatisfiable formula (for a particular number of formal parameters).}. As these clauses are added lazily, we always ask the SAT solver for a new solution and possibly add more \eqref{encode_consistent} if the new solution has a different problem.

\begin{algorithm}
    \caption{Synthesise Substitutions and Effects}
    \label{euclid}
    \begin{algorithmic}[1]
        \For{each distinct action label $l$}
        \State \Call{SynthOne}{$\mathcal{R}_l := \{R \in \mathcal{S} \mid R = (S_b,l,S_a)\}$}
        \EndFor
        \Statex
        \Procedure{SynthOne}{$\mathcal{R}_l$} 
        \State $k \gets \mathit{min\_pars}_l$
        \While{\textbf{not} \Call{SynthOneNumParams}{$\mathcal{R}_l,k$}}
            \State $k \gets k+1$
        \EndWhile
        \State recover and store action $l$ and $\sigma_R$ for every $R \in \mathcal{R}_l$ 
        from the found SAT assignment
        \EndProcedure
        \Statex
        \Procedure{SynthOneNumParams}{$\mathcal{R}_l,k$}
        \State $R \gets \mathcal{R}_l$ \Comment{some transition to start with}
        \Repeat
            \State \Call{EncodeAdditionallyFor}{$R,k$}
            \If{\textbf{not} \Call{Satisfiable}{{}} }
            \State \textbf{return} False
            \EndIf
            \State $R \gets \Call{VerifyAndPickOffending}{\mathcal{R}_l}$
        \Until{$R = \mathit{None}$}
        \State \textbf{return} True
        \EndProcedure
    \end{algorithmic}
\end{algorithm}


Algorithm~\ref{euclid} captures the top-level structure of the just described process.
At the top, we can see that we deal with each action label $l$ in isolation. 
Then just below we fix the speculated number of action parameters $k$ and increment it one by one if necessary. 
Finally, in the last procedure, we highlight the fact that our encoding is incremental, and only if a transition violates a tentative solution (i.e., is \textproc{Offending}) it is encoded into the SAT formula.\footnote{This a key to a good performance on our benchmark traces, as there can be hundreds of transitions sharing the same label, but only just a few needed to synthesise the action that explains everything.} ($\mathit{None}$ is a special value denoting that no such transition exists anymore and thus a solution has been found.) We remark that to save space the actual SAT solver and the encoded formula are only implicit in Algorithm~\ref{euclid} and assumed to be part of a global state.

\subsubsection{Preconditions}

Once the local substitutions $\sigma_R$ are established,
synthesizing preconditions is relatively straightforward
and does not require to be done via SAT encoding.
We strive to generate as many (reasonable) preconditions as possible (while sticking with the empty precondition set would also lead to a solution that explains all the observations) and rely on an idea of ``lifting in all possible ways''. 

For every transition $R=(S_b,l,S_a)$,
we collect all the ground facts in $S_b$ whose arguments are in the range $\sigma_R$. Let us call their set $\mathit{GPC}_R$ for \emph{ground precondition candidates}. (Note that $\mathit{GPC}_R$ includes predicates without arguments, i.e., of zero arity.) Now lifting in all possible ways means forming the set of lifted precondition candidates
\[\mathit{LPC}_R = \{f \mid f\sigma_R\in \mathit{GPC}_R\},\]
where the arguments to $f$ are among the corresponding action established parameters (i.e., the domain of $\sigma_R$).
Finally, we synthesise the precondition set $\mathit{pre}_l$ for the action with label $l$ as the intersection of the candidate sets over all transitions with that label:
\[\mathit{pre}_l = \bigcap_{R \in \mathcal{S}, R = (S_b,l,S_a)} \mathit{LPC}_R.\]
\begin{example}
Lifting in all possible ways is useful for the case of a non-injective $\sigma_R$. Consider, for instance $\sigma_{R} = \{x_1 \mapsto a, x_2 \mapsto a\}$ and $S_b = \{p(a)\}$. Here we speculate $\mathit{LPC}_R = \{p(x_1),p(x_2)\}$, because it is not clear---from this transition $R$---what is the ``right way'' of referring to $p(a)$. This ambiguity typically gets quickly resolved by considering the intersection over all the observed transitions.
\end{example}



\subsubsection{Argument Types}

Finally, we assign types $\vec{t}$ to the formal parameters $\vec{r}$ in the learned actions as follows.
For every object we encounter, as an argument to a (ground) fact in either $S_b$ or $S_a$, we look up the corresponding predicate and the type for the corresponding argument position in its typing information.
For each object, we maintain the most specific (smallest w.r.t.~$\leq$) type obtained this way.
%
%
When it is time to assign types to action parameters, we take the most general type over the objects that ever occurred 
at the particular action argument position. ``Most general'' always makes sense, in the worst case we end up with $\top$.

\section{Limitations of Learning from Logs}

When running a synthesis algorithm on synthetic logs generated from existing domains, it is quite interesting to compare the obtained results with the original domains to see if there are any differences. Here we analyze such differences for the case of running our synthesis algorithm on our IPC-based benchmark. 

\subsubsection{Missing actions}

We obviously cannot re-synthesise actions which do not appear in the input logs. 

\subsubsection{Preconditions}

Our algorithm generally creates more preconditions than just those present in the original problem formulation. One reason for this is that with insufficiently ``exploratory'' logs to synthesise from, the before-states of certain action label's transitions (from which preconditions are synthesised) simply satisfy more properties than they would in general. 

As an example, we can take the IPC benchmark \texttt{Woodworking} and the action \texttt{do-spray-varnish} (with a parameter \texttt{?x - part} among others) for which we synthesise two extra preconditions:
\texttt{(colour ?x natural)} and \texttt{(goalsize ?x small)}. It seems these preconditions only hold by chance in our logs, as nothing should prevent the varnishing actions from spraying on a part that is not of natural color or of other than small size.

A second kind of extra preconditions are those that logically follow from those already stated (within the rules which govern the modelled domain). These include, e.g., \texttt{(road ?l2 ?l1)} in \texttt{Transport}'s \texttt{drive} action, where only \texttt{(road ?l1 ?l2)} is stated normally, but we synthesise both. 
An extreme example of the above phenomenon is preconditions that could be called tautological (from the logical point of view). The \texttt{Agricola} domain contains a rigid predicate \texttt{num\_substract} which tabulates the subtraction relation (i.e., $p1-p2=p3$) for small numbers (type \texttt{num}). This, however, means that when a numerical argument \texttt{?i2} to an action appears, our algorithm proposes \texttt{(num\_substract ?i2 ?i2 num0)} as a precondition, because $x-x=0$ is a mathematical truth. 


There is one exception to the statement that our algorithm only over-approximates the originally stated preconditions. We do not strive to synthesize negative preconditions, and thus, these will always be missing in the synthesised domain. 


\begin{table}[t!]
\renewcommand{\arraystretch}{1.05}
    \centering
    \begin{tabular}{c||r r r |r  r | r r}
Domain & \multicolumn{3}{c|}{count} & \multicolumn{2}{c|}{max. arity} & \multicolumn{2}{c}{traces}\\
Name & \#T & \#P & \#A & $M_P$ & $M_A$ & Tr & Obs\\
\hline
\hline
Barman & 10 & 15 & 10 & 2 & 6 & 4 & 234 \\
Childsnack & 7 & 13 & 6 & 2 & 4 & 4 & 181 \\
Elevators & 6 & 8 & 6 & 2 & 5 & 6 & 142 \\
\hline
Floortile & 4 & 10 & 6 & 2 & 4 & 2 & 80 \\
Hanoi & 2 & 3 & 1 & 2 & 3 & 1 & 7 \\
Nomystery & 6 & 6 & 3 & 3 & 6 & 3 & 41 \\
\hline
Parking & 3 & 5 & 3 & 2 & 3 & 4 & 168 \\
Pegsol & 2 & 5 & 3 & 3 & 3 & 4 & 93 \\
Rovers & 8 & 25 & 9 & 3 & 6 & 3 & 30 \\
\hline
Scanalyzer & 3 & 6 & 4 & 4 & 8 & 4 & 61 \\
Sokoban & 6 & 6 & 3 & 3 & 6 & 4 & 353 \\
Storage & 10 & 8 & 5 & 2 & 5 & 4 & 17 \\
\hline
Termes & 3 & 6 & 7 & 2 & 4 & 4 & 548 \\
Thoughtful & 5 & 18 & 19 & 2 & 7 & 5 & 617 \\
Tidybot & 8 & 24 & 22 & 3 & 9 & 4 & 229 \\
\hline
TPP & 8 & 7 & 4 & 3 & 7 & 4 & 38 \\
Transport & 7 & 5 & 3 & 2 & 5 & 4 & 91 \\
Visitall & 2 & 3 & 1 & 2 & 2 & 4 & 404 \\
    \end{tabular}
    \caption{Details of our benchmark set. The first three columns (count) contain the total number of types (\#T), predicates (\#P), and actions (\#A). We only count actions that occur at least once in the traces used for learning. The next two columns show the maximum parameter arity of the predicates ($M_P$) and actions ($M_A)$. The last two columns
    give the number of traces (Tr) and their total length, i.e., the total
    number of state transition observations (Obs).}
    \label{tab:benchmarks}
\end{table}

\begin{table*}[t!]
\renewcommand{\arraystretch}{1.15}
    \centering
    \begin{tabular}{c||r r r r r | r r r r r| r r r r r}
Domain &  \multicolumn{5}{c|}{L1 Synthesis} & \multicolumn{5}{c|}{SAM} & \multicolumn{5}{c}{Extended SAM}\\
Name & -P & +P & -E & +E & fid. & -P & +P & -E & +E & fid. & -P & +P & -E & +E & fid. \\
\hline
\hline
Barman & 17 & 8 & 0/2 & 0/2 & 0.748 & 0 & 85 & 0 & 0 & 0.835 & 0 & 85 & 0 & 0 & 0.835 \\
Childsnack & 4 & 0 & 0 & 0 & 0.892 & 0 & 22 & 0 & 0 & 0.894 & 0 & 22 & 0 & 0 & 0.894 \\
Elevators & 0 & 10 & 0 & 0 & 0.949 & \multicolumn{5}{c|}{inapplicable} & 0 & 39 & 0 & 0 & 0.826 \\
\hline
Floortile & 3 & 6 & 0 & 0 & 0.893 & 0 & 56 & 0 & 0 & 0.772 & 0 & 56 & 0 & 0 & 0.772 \\
Hanoi & 0 & 1 & 0 & 0 & 0.976 & 0 & 9 & 0 & 0 & 0.816 & 0 & 9 & 0 & 0 & 0.816 \\
Nomystery & 2 & 1 & 0 & 0 & 0.872 & \multicolumn{5}{c|}{inapplicable} & 1 & 9 & 0 & 0 & 0.851 \\
\hline
Parking & 0 & 3 & 0 & 0 & 0.977 & 0 & 19 & 0 & 0 & 0.868 & 0 & 19 & 0 & 0 & 0.868 \\
Pegsol & 1 & 2 & 0 & 0 & 0.952 & 0 & 25 & 0 & 0 & 0.853 & 0 & 25 & 0 & 0 & 0.853 \\
Rovers & 24 & 17 & 12 & 0 & 0.497 & \multicolumn{5}{c|}{inapplicable} & 0 & 105 & 13 & 0 & 0.646 \\
\hline
Scanalyzer & 1 & 7 & 0 & 0 & 0.945 & 0 & 181 & 0 & 0 & 0.537 & 0 & 181 & 0 & 0 & 0.537 \\
Sokoban & 5 & 0 & 0 & 0 & 0.848 & 0 & 25 & 0 & 0 & 0.868 & 0 & 25 & 0 & 0 & 0.868 \\
Storage & 6 & 8 & 0/2 & 0/2 & 0.721 & 0 & 22 & 0 & 0 & 0.896 & 0 & 22 & 0 & 0 & 0.896 \\
\hline
Termes & 21 & 3 & 0 & 0 & 0.546 & 0 & 25 & 0 & 0 & 0.904 & 0 & 25 & 0 & 0 & 0.904 \\
Thoughtful & 35 & 10 & 0 & 0 & 0.854 & 0 & 663 & 0 & 0 & 0.654 & 0 & 665 & 0 & 0 & 0.654 \\
Tidybot & 91 & 37 & 0/4 & 0/4 & 0.556 & \multicolumn{5}{c|}{inapplicable} & 27 & 332 & 11 & 0 & 0.645 \\
\hline
TPP & 8 & 7 & 0/6 & 0/6 & 0.443 & \multicolumn{5}{c|}{inapplicable} & 0 & 54 & 12 & 0 & 0.455 \\
Transport & 0 & 1 & 0 & 0 & 0.990 & 0 & 8 & 0 & 0 & 0.926 & 0 & 8 & 0 & 0 & 0.926 \\
Visitall & 0 & 2 & 0 & 0 & 0.926 & 0 & 3 & 0 & 0 & 0.893 & 0 & 3 & 0 & 0 & 0.893 \\

    \end{tabular}
    \caption{Comparison of the domains produced by the three learning algorithms against the reference (original) domain. We present the total number of missing (-P) and superfluous (+P) preconditions resp. effects (-E and +E) as well as the fidelity value. For effect differences with L1 we present two values in some cases. The first value represents the differences if we tolerate that the parameter types do not match exactly (a sub-type is used instead of the proper type), the second value highlights imperfect matches. More explanation of this subject is provided at the end of the experimental section.
    SAM is 
    inapplicable for the domains 'Elevators', 'Rovers', 'TPP', 'Tidybot', and 'Nomystery'
    since they do not satisfy the injective action binding assumption required by SAM.}
    \label{tab:results}
\end{table*}

\subsubsection{Incorrectly synthesised types}
\label{incorrect-types}
We sometimes synthesise a more general type than the one declared in the original domain. The reason is that the predicates from which we infer the type of an object are typed more generally than the action. As an example, we can name the action \texttt{collect\_animal} from the \texttt{Agricola} domain, where the parameter \texttt{?act} is typed \texttt{animaltag}, but we only infer the more general \texttt{actiontag}, which is how \texttt{?act} is typed in \texttt{available\_action} and \texttt{open\_action} in which it appears there. 

Interestingly, we also observe the opposite, i.e., we synthesise a more specific type than the one mentioned in the original domain presentation. For example, in \texttt{Barman}'s \texttt{clean-shot}, the second argument is originally \texttt{?b - beverage} and the action's body only mentions \texttt{?b} as the second argument of \texttt{used}, also typed as \texttt{beverage} in the domain. Our algorithm, however, synthesised a more specific type \texttt{ingredient} here, because all the invocations of \texttt{clean-shot} in our logs used an object which also appeared as the second argument of \texttt{dispenses}, 
a rigid predicate typing its second argument as \texttt{ingredient}. This could be considered an inefficiency in the \texttt{Barman} domain discovered by our algorithm, as it seems the domain could have typed \texttt{clean-shot}'s second argument \texttt{ingredient} without any loss of generality.

Finally, we may not be able to synthesise any reasonable type for an action parameter, if the corresponding objects in transitions never play a role in any (potential) preconditions or effects. This may sound strange, but happens on several occasions with the \texttt{Woodworking} domain. The gist of the problem is that the domain uses pure object's existence (and none of its properties) as a ``hidden'' precondition of an action. Take, for instance, the action \texttt{do-saw-small} there. There is \texttt{?m - saw} appearing in its parameter list but nowhere else. This (implicitly) means the action is only available in problems where a \texttt{saw} exists. Otherwise, the action cannot be grounded in any way and thus cannot be performed. For obvious reasons, our synthesis has no way of guessing the type of such a parameter. 


\subsubsection{Missing/extra add or delete effects}

Discrepancies in the synthesis of effects again stem from insufficiently diverse observations. We can name the following cases.

With an action in the \texttt{Agricola} domain 
our algorithm failed to synthesise a few effects such as \texttt{(available\_action act\_sow)}. The reason for this could be that the domain actually does not have an action by which \texttt{(available\_action act\_sow)} could be deleted and so adding it is always a moot effect (provided initial states contain it, which they in our case do).

With the \texttt{Rovers} domain there are several actions \texttt{communicate\_<something>\_data}, which at the same time both add and delete \texttt{(channel\_free ?l)} and \texttt{(available ?r)}. This essentially boils down just to adding these facts (as additions have precedence over deletions), but in our traces the corresponding facts are always true and so the effects become invisible for our synthesis\footnote{This seems to be 
an artifact from translating the domain from a more expressive formalism.}. 

Finally, we observed two cases in which a set of effects got synthesised, but with slight discrepancies in the parameter usage. In \texttt{Tidybot}'s \texttt{base-cart-right}, the action comes with a long list of parameters 
and our algorithm synthesised, e.g., \texttt{(base-obstacle ?x2 ?cy)} instead of \texttt{(base-obstacle ?cx2 ?cy)}. 
This can be explained by noticing that our logs did not capture the full generality of instantiating all the parameters with distinct values. For example, \texttt{?x2} was always instantiated with the same objects as \texttt{?cx2}, 
thus our algorithm could not distinguish the two and picked the ``wrong'' version of the effect. 


\section{Experimental Evaluation}
We compare our new algorithm (called L1) against the SAM and Extended SAM (E-SAM) algorithms~\cite{sam}. To our best knowledge, these algorithms are the most recent and most similar to our proposed algorithm. All three algorithms generate STRIPS models from precise traces (no missing or noisy information is allowed). As for the differences, SAM and E-SAM are given the list of all action parameters (including their types) in the traces and also in the lifted actions to be learned as input. Therefore SAM and E-SAM only need to figure out the preconditions and effects of each lifted action. On the other hand, L1 does not have access to this information and needs to infer it. Furthermore, SAM has an additional requirement on the traces called the injective action binding assumption, which makes it inapplicable for some domains.

The implementation of our L1 algorithm is done in Python using the pysat~\cite{Stoneback2018} library for SAT solving and Glucose~3.0~\cite{audemard:hal-03299473} as the SAT solver. The evaluation was run on a computer with an Intel(R) Core(TM) i7-12700H 2.30 GHz CPU and 64 GB of RAM. The time limit for the learning of the domain models was set to 1 minute.

We evaluate the algorithms by comparing the learned actions to the original actions from the reference domains, i.e., the domains used to generate the benchmark traces. Note that this is a strong requirement for the equivalence of domain models as, technically speaking, models can be equivalent (i.e., generating the same plans) even with different action schemas. Yet, such a domain equivalence might be even undecidable. 

The benchmark domains were selected from the domains used in recent International Planning Competitions~\cite{vallati20152014}. We filtered out domains that did not use types and featured non-strips constructs such as conditional effects, quantifiers and numeric operations.
The details of all the benchmark domains and their corresponding traces are given in Table~\ref{tab:benchmarks}. The reference domains, traces, and learned domains by our algorithm are available online at \url{https://github.com/FilutaAI/synthesis-benchmarks}. 

The results are summarized in Table~\ref{tab:results}. We present the total number of missing (-P) and superfluous (+P) preconditions resp.~effects (-E and +E) in the learned domains.
We also use a composite measure of performance, which we call \emph{fidelity}. 
It is defined as
\[
\mathrm{fid.}=\frac{\mathrm{mapped}\_\mathrm{predicates}}{\mathrm{mapped}\_\mathrm{predicates}+\mathrm{score}}\;,
\]
where $\mathrm{mapped}\_\mathrm{predicates}$ is the sum of the numbers of matched preconditions and effects, and $\mathrm{score}=(\text{-P})+0.2(\text{+P})+(\text{-E})+(\text{+E})$ is a weighted sum of unmatched predicates. The extra preconditions are weighted by factor $0.2$\footnote{The value $0.2$ was chosen arbitrarily.}, 
since we prefer rather to overestimate than underestimate the set of preconditions. The rationale for this is that it is easier for a human modeler to remove unnecessary preconditions than to think of and add missing ones. Others are weighted with factor one, which puts them on the same footing as the mapped predicates. This measure takes a value between zero and one and can be interpreted as a percentage of how well the synthesized model reproduces the original.

From the results in Table~\ref{tab:results} we can make several observations. Firstly, SAM is inapplicable for 5 of the 18 domains since these domains do not satisfy the injective action binding assumption. Next, both SAM algorithms tend to have a very high number of superfluous preconditions, significantly higher than L1 for most domains. On the other hand, they never learn superfluous effects. In a sense, neither does L1, unless we are doing a strict comparison where we do not tolerate different types. The indicated superfluous effects learned by L1 are due to learning the type of the parameter incorrectly, we will explain this in more detail below.

As one would expect, the performance of E-SAM is very similar to SAM and on domains satisfying the injective action binding assumption.

Looking at the results of our L1 algorithm, we can see that the learned domains are mostly  close to the corresponding reference domains, most differing due to missing and superfluous preconditions. The lowest fidelity scores are present in the Rovers, Termes, Tidybot, and TPP domains. Interestingly, three of these domains do not satisfy the injective binding assumption, which appears to be an indication of their difficulty. On the other hand, L1 works very well on Elevators and Nomystery, which also do not satisfy the assumption.

Regarding runtime, we do not present detailed results and only summarize that all three algorithms are very fast on the used benchmarks taking only a few seconds to complete all the tests. 

\subsection{The superfluous effects of L1}
The superfluous effects learned by L1 are actually correct but with a wrong parameter type. Let us clarify this on the example of the Barman domain. Barman contains the types \texttt{shot}, \texttt{beverage} and its sub-type \texttt{ingredient}. There exists an action \texttt{clean-shot} with the parameters \texttt{(?s - shot ?b - beverage)} and an effect \texttt{(not(used ?s ?b))}. However, L1 synthesis learns \texttt{clean-shot} with parameters \texttt{(?s - shot ?i - ingredient)} and hence with the effect \texttt{(not(used~?s~?i))}. This is evaluated as one missing and one superfluous effect. All of the superfluous effects in Table~\ref{tab:results} are of this kind, which is also the reason for there being at least as many missing effects in each case. For further reading, we refer to the `incorrectly synthesized types' paragraph in the `limitations of learning from logs' Section. 

Also note that this kind of mistake is not possible in SAM and E-SAM, because the list of action parameters with their types is given as input to them.

\section{Conclusion}
In this paper, we introduced a novel algorithm, called L1, that learns STRIPS action models from fully observable state traces that do not contain information about the parameters of the actions (in contrast to existing similar algorithms). We have shown that action synthesis with a bounded number of parameters is at least as hard as the graph isomorphism problem, NP-hardness remains an open problem. The L1 algorithm is based on providing \emph{explanations} for transitions between the states in the state traces. That means, in a nutshell, that the synthesized action schema has to correctly ``fit'' all the transitions in which the action is involved, i.e., for a transition $S,a,S'$, $\mathit{pre}(a)\subseteq S$ and $S'=\mathit{apply}(S,a)$. L1 encodes the problem of action model synthesis into Boolean Satisfiability (SAT) as it is an effective way to deal with possible backtracking that might occur during the synthesis process.

We have empirically shown that our L1 algorithm is capable of learning more accurate action models than the state-of-the-art algorithms SAM and Extended SAM in multiple domains despite the fact they use more information about the actions to be learned and strictly more information in the traces.

As for future work, we plan to develop an algorithm called L0, that would require even less information in the state traces. In contrast to L1, L0 traces will not require action labels, i.e., it will synthesise action models only from information about state trajectories. Furthermore, we plan to consider ADL features such as conditional effects that allow synthesizing more compact (and practical) action models.

\bibliographystyle{kr}
\bibliography{LaTeX/refs}

\end{document}